\documentclass{article}
\begin{document}

\subsection*{Review of: Brigitte Le Roux and Henry Rouanet, 
Geometric Data Analysis, From Correspondence Analysis to Structured 
Data Analysis, Kluwer, Dordrecht, 2004, xi+475 pp.}

The term ``Geometric Data Analysis'' is due to Patrick Suppes 
(Stanford) who writes a Foreword for this encyclopedic view of 
Correspondence Analysis.
The uniqueness of this work lies in the detailed conceptual 
framework, and in showing how, where and why statistical inference
methods come into play.  
  
As a data analysis methodology, 
Correspondence Analysis is {\em formal} and {\em geometric}.
The former follows from the fact that mathematical ``structures govern
procedures'', 
the mathematics carry the burden of demonstration, and ultimately 
``this means a fantastic saving in intellectual investment''.  
The latter is due to the focus on clouds
of points in geometric spaces.  Unlike the sampling-oriented
approach of a good deal of statistical data analysis, including
multivariate data analysis, for Correspondence Analysis, ``description 
comes first'' since statistics is not  reducible to probability.  

The word ``model'' is used, in general, in lots and lots of 
senses (statistical, mathematical, physical models; mixture 
model; linear model; noise model; 
neural network model; sparse decomposition model; even data model).  So 
it is helpful that the methodological models under discussion in this 
book are termed {frame models}.  As a {\em geometric frame model}, it 
is also nice to see due emphasis given to ``Euclidean classification'' as 
a ``companion method'' of Correspondence Analysis.  
Euclidean classification is most in harmony with Correspondence Analysis
when minimum variance (or Ward's) agglomerative hierarchical clustering is 
used.   The data aggregates are defined in the same 
Correspondence Analysis metric spaces.  The clustering is Euclidean 
because it is constructed in the 
Correspondence Analysis factor space. 

In social sciences, there are ``two distinct conceptions of the role
of statistics, namely sustaining a {\em sociology of variables} versus
{\em constructing a social space}''.  
 The former goes hand in 
hand with traditional statistical analysis, whereas the latter is 
part and parcel of Geometric Data Analysis.  We will return to the 
construction of social space below with reference to Bourdieu.

As a ``cornerstone'' of Correspondence Analysis, one has the 
``measure versus variable duality'', or the duality of the 
associated (row, column) vector spaces.  When aggregated,
the measures are summed, whereas the variables,
when they are aggregated, are averaged.  A transition notation 
where subscripts denote measures and superscripts denote variables 
is handy -- much handier than matrix notation -- for important 
aspects of the analysis.  Among the latter are: 
how one's data departs from the 
``reference''  marginal frequencies; how one's data is expressed as
 a ``transition'' from one to the other of these marginal frequencies; 
and how the $\chi^2$ metric is defined on the dual, complementary clouds.

Discussing the suitability of Correspondence Analysis for structured
data (that is to say, structured around such variables as
age and gender characteristics in a questionnaire), 
firstly having such structuring factors as supplementary variables 
allows for their incorporation into the analysis.  Later, other 
analyses such as Analysis of Variance (ANOVA) could be very appropriate.
Statistical inference is a natural complement to descriptive 
analysis -- as Benz\'ecri wrote, 
``The model should follow the data, not the reverse!'' -- and both 
phases together comprise a powerful way to inductively carry out  
data analysis.  

Chapter 2 presents a most accessible and readable 
introduction to Correspondence Analysis,
based on the view of the measure versus variable duality.  This leads 
to Correspondence Analysis being viewed 
as analysis of a measure over the Cartesian
product with strictly positive marginals.  
All of the essentials of Correspondence Analysis are introduced, going as
far as aids to interpretation (e.g.\ contributions of points 
to the inertia of axes; quality of representation; 
data coding; supplementary elements; 
linkages with matrix expression, Fisher's linear discriminant analysis, 
regression, and canonical correlation analysis; links with multidimensional
scaling; and how or where CA links up with probabilistic model frames).
A range of questions end the chapter, with solutions and with comments.  

Chapter 3 dealing with analysis of a Euclidean cloud of points proceeds 
by way of Huyghens, and spectral decomposition (i.e., eigenvalue, eigenvector
decomposition) to concentration (hyper)ellipsoids, and following 
discussion of the partition of a cloud the reader is led to clustering.
Pride of place is accorded agglomerative hierarchical clustering with the 
minimum variance criterion (Ward's method; ``Euclidean classification'' 
since the input consists of Euclidean projections on factors).  The 
reciprocal nearest neighbors algorithm is described but not the
nearest neighbor chain algorithm  
(which is 
more manageable from a computational complexity point of view, since 
it has an $O(n^2)$ computational bound for most widely used agglomerative 
criteria). For the nearest neighbor chain and reciprocal nearest neighbor
algorithms, which came out of Benz\'ecri's lab in the early 
1980s and are now in R and Clustan among other packages, see Murtagh (2005).  
The ultrametric is not discussed.  For the view of 
Correspondence Analysis being a ``tale of three metrics'', $\chi^2$,
Euclidean and ultrametric, see Murtagh (2005).  
Chapter 3 ends with exercises and solutions.  

Chapter 4 deals comprehensively with Principal Components Analysis, viewed 
from the points of view of formal description and also interpretation of 
results, through the prism of Correspondence Analysis.  In terms of number 
of pages it is more than half the length of Dunteman (1989) (but
admittedly shorter 
than the 500 page length Jolliffe, 2002).  The treatment of 
PCA in Le Roux and Rouanet is unique and in the spirit of 
Geometrical Data Analysis.  

In chapter 5, attention is given to Multiple Correspondence Analysis.  
A paradigmatic case is questionnaire analysis, where each question has a 
number of response modalities, and just one among these response modalities
is indicated by a subject.  Much that is theoretical (e.g., Burt table; 
viewpoints linked to discriminant analysis, and canonical correlation 
analysis) and practical (e.g., choice of active questions; data coding
for handling infrequent modalities) is covered.  Before ending 
the chapter with 
questions and solutions, there is a detailed case study, relating to a 
French Government survey of cultural and leisure activities (3002 respondents,
carried out in 1997).  

We have already mentioned ``structured data'' above.  Examples 
of such structuring context, available in addition to the essential data being
studied, include gender, age and education variables in the 
cultural and leisure 
questionnaire; or authors and years in textual-based authorship assessment.
Due to the overall aim of studying the effect of experimental 
(independent, under analyst control) variables on the dependent 
variables (response variables), and also the need often to take 
nesting and crossing structures of variables into account, there is 
linkage in chapter 6, ``Structured Data Analysis'', with ANOVA, MANOVA
and regression.  These techniques are discussed within the model frame 
of Correspondence Analysis: they are ``grafted'' 
into the Geometric Data Analysis framework.  A case study is provided based on 
visually-based annotated video of basketball players, with the aim of 
selecting potential high achievers.  

Chapter 7 on stability analysis first touches on alternatives, functional
analysis (Krzanowski, Critchley, Tanaka, Pack and Jolliffe, B\'enass\'eni),
and bootstrapping (Diaconis and Efron, Daudin et al., Lebart et al.), before
returning to the approach of Escofier and Le Roux that perturbs a cloud
relative to a reference cloud, and checks what are the effects of doing this.
A range of case studies is used to investigate an outcome
without a given group being present, or to investigate an outcome following 
deletion of a single point, or to investigate the discarding of variables
or modalities.  A range of stability results for eigenvalues is also 
considered.

Often in applications inference is needed, based on statistical modeling,
and the authors encourage a comprehensive approach to ``Inductive Data 
Analysis'' in chapter 8.  The point is made that descriptive statistics 
are based on relative frequencies and do not depend on sample size, 
whereas test statistics (for induction or inference) do combine an erstwhile
descriptive statistic with the sample size.  There is much to reflect on 
in this chapter.  Traditional significance testing, frequentist and 
combinatorial inference, Bayesian inference, -- all are discussed in a 
way that is well-grounded geometrically and eminently readable.  

Chapter 9, 85 pages in length, deals with three major case studies: a 
medication treatment study of Parkinson's patients; French political 
attitudes before, and voting in, elections in 1997; and part of a large
``Education Program for Gifted Youth (EPGY)'' study carried out at 
Stanford University.  There is a great wealth of experience etched 
deeply into these application studies.  

An overview of applicable mathematics rounds off the book.  We have 
noted earlier the authors' 
eschewing of matrix formalism in favor of 
transitions.  Matrix description is to be found throughout, but the 
appropriateness and powerfulness of the transition based approach 
is given pride of place.
The transition formalism uses quite lightly the Einstein tensor notation 
that is used more extensively in 
Benz\'ecri (1973).  Van Rijsbergen (2004) provides 
another very commendable survey of geometric data analysis in information 
retrieval, using Dirac notation that is now standard in quantum physics.

The transition formalism is nicely expressed by mapping, or morphism, 
diagrams (as were used a great deal in Cailliez and Pag\`es, 1976; and which 
have been taken further in McCullagh, 2002, who appraises statistical 
modeling through the prism of category theory).  Le Roux and Rouanet are
unique in their succinct introduction to homomorphisms and 
endomorphisms, the application of spectral decomposition in this context, and
discussion of extremality properties.  

A major feature of this book, following the example of Benz\'ecri, 
is the close practical and experienced 
attention throughout paid to the epistemology of data 
analysis, i.e.\ the interface between data and reality.  This is 
nowhere stronger than in the authors' continuation 
of Bourdieu's 
(see e.g.\ Bourdieu, 1984) approach to mapping out of a social field
or space.  In chapters 5, 6 and 9, there is appreciable discussion of
Bourdieu's use of Geometric Data Analysis. 

In conclusion, this book constitutes essential background material on 
Geometric Data Analysis, and, for the seasoned professional, a most
valuable source of reference.  

\subsection*{References}

\medskip
\noindent
BENZ\'ECRI, J.P. et Coll. (1973). {\em 
L'Analyse des Donn\'ees. Vol. 1: 
Taxinomie.  Vol. 2: Correspondances},  Paris: Dunod.  (1976, 2nd edn.)

\medskip
\noindent
BOURDIEU, P. 
(1984).  {\em Distinction: A Social Critique of the Judgement of Taste},
translated by R. Nice. 
Cambridge, MA: Harvard University Press.  (1979) 
{\em La Distinction: Critique Sociale du Jugement}, 
Paris: Les \'Editions de Minuit.

\medskip
\noindent
CAILLIEZ, F. and PAG\`ES, J.-P. (1976). 
{\em Introduction \`a l'Analyse des 
Donn\'ees}. Paris: Soci\'et\'e de 
Math\'ematiques Appliqu\'ees et de Sciences Humaines. SMASH.

\medskip
\noindent
DUNTEMAN, G.H. (1989). 
{\em Principal Components Analysis}.  Thousand Oaks, CA: Sage Publications.

\medskip
\noindent
JOLLIFFE, I.T. (2002).
{\em Principal Components Analysis}, 2nd ed., New York: Springer. 

\medskip
\noindent
McCULLAGH, P. (2002).
``What is a statistical model?'', {\em Annals of Statistics}, 
30, 1225--1310.  

\medskip
\noindent
MURTAGH, F. (2005). 
{\em Correspondence Analysis and Data Coding with Java and R},
Boca Raton, FL: Chapman and Hall/CRC.

\medskip
\noindent
VAN RIJSBERGEN, K. (2004). 
{\em The Geometry of Information Retrieval}, 
Cambridge: Cambridge University Press.

\bigskip

\bigskip

\noindent
Fionn Murtagh

\noindent
Science Foundation Ireland 

\noindent
and 

\noindent
Department of Computer Science, Royal Holloway, University of London

\end{document}